\documentclass{bmvc2k}

\usepackage{graphicx}
\usepackage{booktabs}
\usepackage[accsupp]{axessibility}
\usepackage{orcidlink}
\usepackage[most]{tcolorbox}
\usepackage{mathptmx} 
\usepackage[font=footnotesize,skip=1pt]{caption}
\usepackage[dvipsnames]{xcolor}
\usepackage[ruled,vlined]{algorithm2e}
\usepackage{tabularx} 
\usepackage{pifont}
\usepackage{multirow}
\usepackage{graphicx}
\usepackage{float}
\usepackage{hyperref}
\usepackage{titletoc}
\usepackage{wrapfig}
\usepackage{arydshln}
\usepackage{cleveref}
\usepackage{lineno}
\definecolor{lightgrayrow}{gray}{0.94}

\newcommand{\modelname}{BiScopeNet\xspace}
\newcommand{\datasetname}{BiScopeCap\xspace}
\newcommand{\projectionmodule}{FocusAlign\xspace}
\newcommand{\newloss}{MetaLoss\xspace}

\newcommand{\cmark}{\ding{51}}
\newcommand{\xmark}{\ding{55}}

\newcounter{inlinesubsec}

\title{Sub-actions in Action: Text-Guided Hand-Role Alignment for Sub-Action Recognition}

\addauthor{Sadegh Rahmaniboldaji}{s.rahmani@surrey.ac.uk}{1}
\addauthor{Filip Rybansky}{f.rybansky2@newcastle.ac.uk}{2}
\addauthor{Quoc Vuong}{quoc.vuong@newcastle.ac.uk}{2}
\addauthor{Frank Guerin}{f.guerin@surrey.ac.uk}{1}
\addauthor{Andrew Gilbert}{a.gilbert@surrey.ac.uk}{1}

\addinstitution{
 University of Surrey\\
 Surrey, UK
}
\addinstitution{
 University of Newcastle\\
 Newcastle, UK
}

\runninghead{Rahmani ET AL.}{Sub-actions in action}

\begin{document}

\maketitle

\begin{abstract}
Our objective is to enhance VLMs' ability to distinguish the roles of the left and right hands in egocentric actions. We introduce \textbf{sub-action detection}, a task that requires models to explicitly recognise the left‑ and right‑hand roles and their manipulated objects within an ongoing high‑level action. To address this challenge, we propose \textbf{\modelname}, a framework that augments existing video‑LLMs without modifying their vision encoders or language models. \modelname incorporates two components: \textbf{\projectionmodule}, a gated text-guided 
attention-based module that selectively aligns video features with textual context to emphasise action-relevant cues, and \textbf{\newloss}, a masked, object-weighted loss that prioritises visually grounded terms.
To supervise sub‑action detection, we introduce \textbf{\datasetname-100K}, a large-scale video-level dataset curated from existing egocentric datasets, providing dense captions that jointly describe high‑level actions, left/right‑hand roles, and manipulated objects, enriched with spatial metadata for grounding. Across the evaluated Video-LLM backbones, \modelname improves action, hand-role, and object recognition over the corresponding supervised baselines, achieving the strongest aggregate performance among the evaluated configurations on \datasetname. The project page is available at: \url{https://andrewjohngilbert.github.io/BiScopeNet/}

\end{abstract}

\section{Introduction}
\label{sec:intro}

Online video content and embodied agents increasingly demand structured video understanding beyond coarse action labels, as real-world manipulation tasks often involve multiple sub-actions. However, in many action recognition approaches, a caption typically describes the high-level action (e.g., “a person is picking up a pan”). In contrast, sub-actions capture the fine-grained hand-specific contributions within the same activity (e.g. “a person is picking up a pan. The right hand holds the pan and the left hand holds the lid”). Capturing such fine-grained hand-role semantics is important for applications that require precise interaction understanding, including human–robot collaboration, AR/VR environments, and educational or professional training scenarios, where identifying each hand's contribution can improve task interpretation and responsiveness.

Recent advances in video descriptors \cite{Maaz2024VideoGPT+, damonlpsg2024videollama2, marafioti2025smolvlm, Maaz2023VideoChatGPT, nvila2025, qwen2025qwen35, bai2025qwen25vl, yang2024qwen2technicalreport, chen2024internvl} have significantly improved video understanding on various large-scale benchmarks \cite{ego4d, egoexo4d, epic_Damen2018EPICKITCHENS}. However, progress is still primarily evaluated at the level of high-level action recognition or captioning. As a result, models often fail to distinguish hand-specific sub-actions, such as determining which hand manipulates an object and which provides support during an ongoing activity. In egocentric settings, motion blur, occlusion, and visual clutter further complicate the extraction of such fine-grained cues. In this work, we explicitly model sub-actions at the hand level by introducing a structured framework that differentiates between left- and right-hand roles within ongoing high-level actions.

\begin{figure}[t]
\centering
\includegraphics[width=\columnwidth]{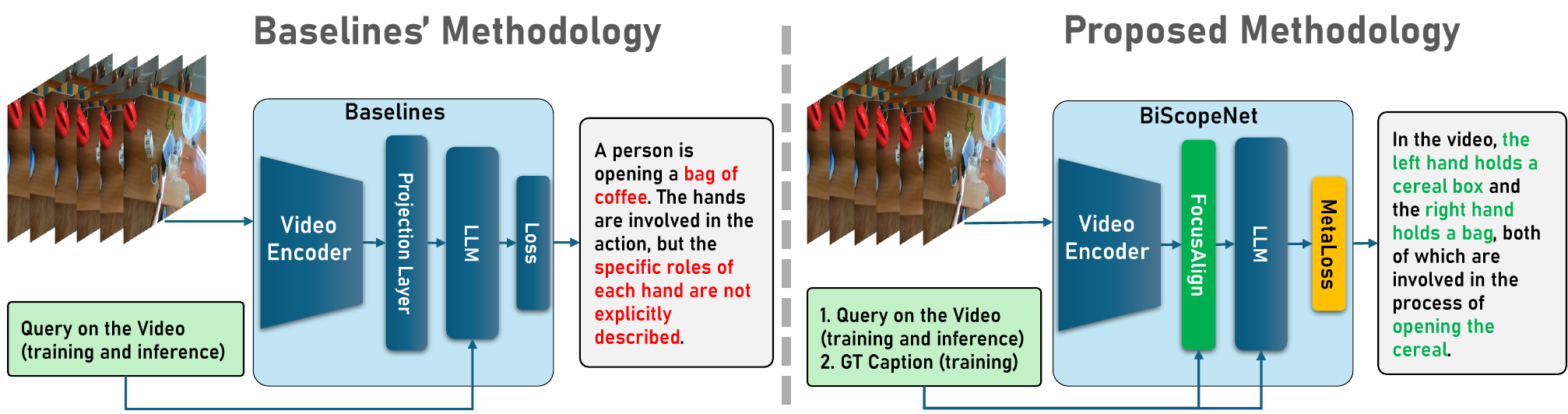}
\caption{Comparison of \modelname and baseline models, showing how the new data incorporation approach, along with the proposed projection and loss mechanisms, improves hand sub-action recognition, overall action understanding, and object grounding in generated captions.}
\label{fig:biscopenet_vs_baselines}
\end{figure}

To enable sub-action detection of fine-grained videos, we propose \textbf{\modelname}, which consists of two components: \textbf{\projectionmodule}, a text-guided projection module, and \textbf{\newloss}, a selective training loss. \projectionmodule refines the visual representation before it reaches the language model. Conventional linear or MLP-based projection layers \cite{Maaz2023VideoChatGPT, Maaz2024VideoGPT+} treat visual features uniformly and limit fine-grained alignment with textual cues. Query-based methods such as Q-Former~\cite{li2023blip2} and Flamingo’s perceiver resampler~\cite{NEURIPS2022_960a172b} compress visual tokens into a fixed set of latent queries prior to cross-modal fusion. Unlike these approaches, our design introduces a structure with visual and textual information exchanged via an attention mechanism and refined via an adaptive, gated residual connection, passing features through an auxiliary alignment MLP. Rather than relying on a single interaction mechanism, our design jointly models cross-modal interactions while preserving text as the guiding signal and dynamically reweighting visual representations. This combination enables the model to prioritise action- and sub-action-relevant visual features, maintaining robust alignment even during inference when textual input is minimal.

In parallel, we introduced \newloss, a masked and object-weighted loss that ignores spatial metadata tokens used for structural guidance while increasing the contribution of grounded object tokens. Unlike standard cross-entropy loss, which treats all tokens equally, this selective formulation promotes stronger visual grounding in generated captions. Prior selective objectives such as RHO-LOSS~\cite{pmlr-v162-mindermann22a_rho_loss} and RHO-1~\cite{NEURIPS2024_3322a9a7rho1} rely on auxiliary models or adaptive scoring mechanisms to prioritise examples or tokens during training. In contrast, \newloss derives masking decisions deterministically from annotation structure, requiring no auxiliary reference model, while additionally introducing an explicit object-weighting term ($\lambda_o = 3$) that amplifies gradient signals on visually grounded object tokens, a mechanism absent from prior selective training formulations. \Cref{fig:biscopenet_vs_baselines} contrasts our approach with baseline architectures, and detailed ablations are provided in \Cref{sec:experiments}.

Effective learning also requires supervision that reflects hand-specific structure. Existing egocentric datasets only partially satisfy this requirement. EPIC-KITCHENS/VISOR \cite{epic_Damen2018EPICKITCHENS, epic_VISOR2022} and Ego4D \cite{ego4d} provide large-scale coverage but focus primarily on high-level action labels without explicit hand-role supervision. HOI-QA \cite{hoiqa_bansal2024hoi} includes sparse hand annotations limited to selected frames, and HD-Epic \cite{perrett2025hdepic} offers richer captions yet lacks explicit modelling of both high-level action labels and per-hand roles within unified video-level descriptions. Consequently, current benchmarks do not adequately support joint reasoning over high-level actions and hand-specific sub-actions.

To establish a sub-action evaluation benchmark, we introduce \textbf{\datasetname-100K}, a video-level corpus that provides dense captions describing high-level actions, along with explicit left- and right-hand roles, and is augmented with spatial metadata for grounding. The dataset serves as structured supervision for action- and sub-action detection. A detailed comparison with existing datasets is provided in \Cref{tab:dataset_full_comparison}.

Our main contributions are summarised as follows:

\begin{enumerate}

\item \textbf{Sub-action Detection Task:} We introduce a new task, \textbf{sub-action detection}, which requires models to explicitly recognise left- and right-hand roles within an ongoing action, alongside the high-level action itself.

\item \textbf{\modelname:} We propose a framework composed of two components: \textbf{\projectionmodule}, a text-guided module that aligns video features with textual context to emphasise action and sub-action cues, and \textbf{\newloss}, a masked, object-weighted objective that suppresses structural tokens and strengthens visual grounding for fine-grained captioning.

\item \textbf{Dataset and SOTA results:} We introduce \textbf{\datasetname-100K}, a large-scale video-level benchmark with dense captions covering both global actions and fine-grained hand-specific sub-actions, enriched with spatial metadata for structured supervision. Using this fine-grained supervision, \modelname outperforms recent video descriptors.

\end{enumerate}
  
\section{Related Work}
\label{sec:relatework}

This section reviews prior work relevant to fine-grained egocentric action captioning. We first discuss foundational video modelling and the evolution from action recognition to video captioning, then examine approaches for egocentric action understanding. Finally, we review visual–language fusion strategies in Video-LLMs and recent efforts toward semantic evaluation using LLM-based judges.

\noindent\textbf{Foundations of Video Modelling.}
Early video understanding established volumetric spatiotemporal feature learning with 3D CNNs such as C3D and I3D \cite{Learning_Spatiotemporal_Features_with_3D_CNN_c3d, carreira2017Quo_vadis_i3d}. Later work improved long-range temporal reasoning via non-local operations \cite{Wang_2018_CVPR_Non-Local_Neural_Networks} and dual-pathway architectures such as SlowFast \cite{Feichtenhofer_2019_ICCV_SlowFast}, while efficient scaling strategies such as X3D \cite{feichtenhofer2020x3d} balanced speed and accuracy. Transformer-based models, including TimeSformer \cite{gberta_2021_ICML_timesformer}, ViViT \cite{Arnab_2021_ICCV_ViViT}, and VidTr \cite{zhang2021vidtr}, introduced global self-attention, enhancing the modelling of short, rapid motions critical for egocentric hand manipulations. Despite these advances, most approaches focus on clip-level action recognition rather than fine-grained, language-grounded descriptions.

\noindent\textbf{From Action Recognition to Video Captioning.}

Video captioning extends action recognition by generating temporally grounded natural language. Many standard captioning datasets lack the fine-grained, manipulation-centred annotations needed for egocentric sub-action narratives \cite{krishna2017dense_activityNet_captions, 10.5555/3504035.3504965_youcook2, sigurdsson2018charadesegolargescaledatasetpaired, epic_Damen2018EPICKITCHENS, ego4d}. Recent Video-LLMs, including Momentor \cite{qian2024momentor}, TimeChat \cite{ren2024timechat}, VTG-LLM \cite{guo2025vtg}, AVicuna \cite{tang2025empowering_avicuna}, and  Vid2Seq \cite{yang2023vid2seq} leverage pretrained LLMs to generate coherent narratives with improved temporal reasoning.
However, the strong linguistic capabilities of LLM-based approaches do not necessarily guarantee faithful grounding in subtle visual dynamics, particularly for fine-grained interactions \cite{3b62dba1310749008ba2ab491a6e6c26}.

\noindent\textbf{Egocentric Action Understanding and Manipulation Captioning.}

Egocentric video understanding presents unique challenges, including limited field of view, frequent hand–object occlusions, and rapid interactions. Exo2EgoDVC \cite{ohkawa:wacv25_exo2egoDVC} addresses limited labelled data by aligning egocentric cooking videos with YouCook2, while EgoVideo \cite{pei2024egovideoexploringegocentricfoundation} emphasises modelling fine-grained hand–object state transitions beyond high-level action labels, focusing on scene and background details rather than on sub-actions. Motion-aware captioning methods such as the Motion-Augmented Caption Model (M-ACM) \cite{10.1145/3746027.3754560} further improve motion-sensitive descriptions using dual-pathway streams and human mesh recovery. Despite these advances, most approaches operate at event level or generate high-level narratives. Fine-grained sub-action detection, capturing subtle hand roles, contact states, and short temporal transitions, remains underexplored in captioning architectures, despite its importance for precise manipulation semantics.

\noindent\textbf{Visual–Language Fusion in Video-LLMs.}

A key design choice in Video-LLMs is how visual features are integrated into language models \cite{tang2025videounderstandinglargelanguage}, either via Analyser → LLM pipelines \cite{li2024videochatchatcentricvideounderstanding} that convert videos into intermediate text/structures, or Embedder → LLM that project visual features directly into the LLM token space. Within the latter, prior work uses linear or MLP projectors \cite{Maaz2023VideoChatGPT, Maaz2024VideoGPT+, marafioti2025smolvlm, zhou2024tinyllava, li2024llava_onevision, li2024groundinggpt} or Q-Former \cite{li2023blip2}. Q-Former compresses visual content into query tokens, potentially discarding fine-grained spatiotemporal cues. Separately, Flamingo \cite{NEURIPS2022_960a172b} compresses visual tokens into a fixed set of latent queries via a perceiver resampler and introduces gated cross-attention layers interleaved within the LLM decoder. While effective for general vision-language tasks, the perceiver compression reduces spatial granularity before cross-modal fusion begins, potentially discarding the fine-grained, localised cues that hand-role disambiguation requires 
In contrast, our proposed projection layer, \projectionmodule, operates on the full spatiotemporal grid with caption-conditioned gating to suppress irrelevant background regions, preserving fine-grained hand–object dynamics without aggressive compression.

\noindent\textbf{Semantic Evaluation with LLM Judges.}
As captioning becomes more fine-grained, traditional n-gram metrics and evaluation protocols such as CIDEr \cite{Vedantam_2015_CVPR_cider}, METEOR \cite{10.5555/1626355.1626389_Meteor}, and SODA/F1 \cite{10.1007/978-3-030-58539-6_31_soda} struggle to capture semantic faithfulness and temporal consistency. Recent studies show that LLM-based judges provide scalable evaluation aligned with human reasoning. A large-scale study in the TREC 2024 RAG Track found that LLMs achieved high correlation with human assessments and perfect agreement on support evaluation cases \cite{thakur2025supportevaluationtrec2024}. Similarly, VideoAutoArena and ARGUS use GPT-4o to detect hallucinations and temporal inconsistencies in Video-LLMs \cite{11094731, Rawal_2025_ICCv}. Earlier works such as \cite{Maaz2023VideoChatGPT} also employed LLMs in their evaluation pipelines. Motivated by these findings, we adopt an LLM-as-a-judge framework to assess semantic fidelity in egocentric action/sub-action captioning.
\label{sec:methodology}
\section{Methodology}

\begin{figure*}[!t] 
\centering
\includegraphics[width=\linewidth]{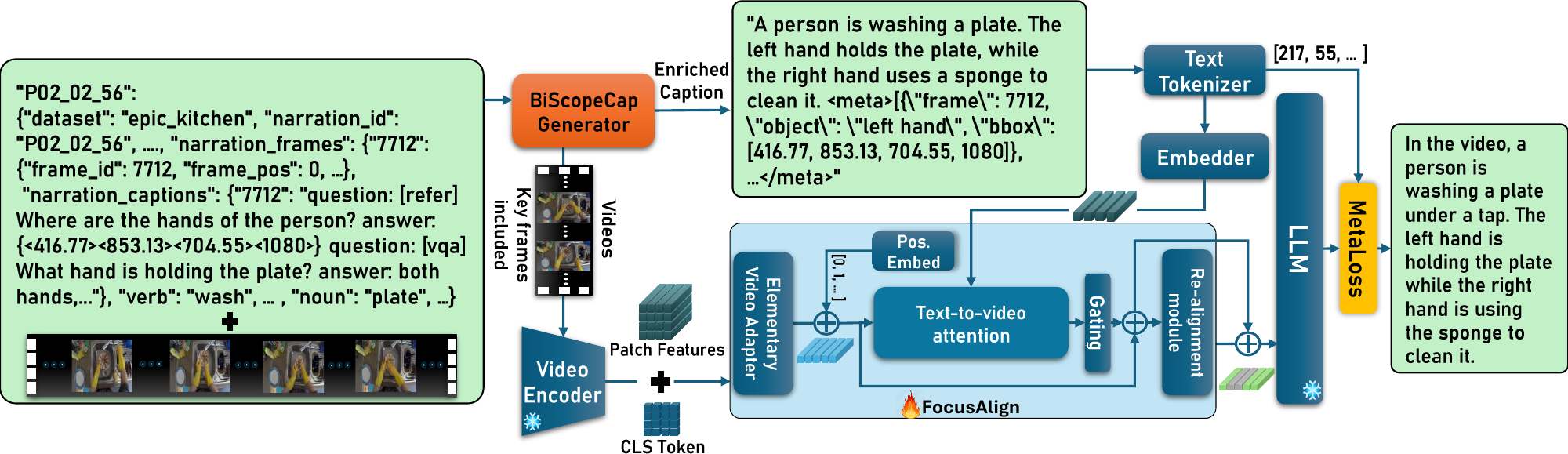}
\caption{\textbf{Overview of BiScopeNet.} Video frames are encoded into visual tokens and processed by our \projectionmodule, which integrates text-guided cross-attention, adaptive gating, and MLP refinement to prioritise action-relevant patches. The resulting representation is projected into the LLM space to generate fine-grained captions that capture hand interactions underlying the main action.}
\label{fig:model_architecture}
\end{figure*}

We present \modelname, a framework for \emph{sub-action detection} in egocentric video captioning. 
The design of \modelname is driven by two principles:  (i) sub-action aware cross-modal alignment, and (ii) selective supervision that prioritises visually grounded hand–object cues. We first describe the overall architecture, then detail the two key components: \projectionmodule, which performs task-conditioned cross-modal feature alignment, and \newloss, which enforces structured and object-aware supervision.

\subsection{\modelname Framework}

\modelname is a backbone-agnostic multimodal captioning framework designed specifically for sub-action detection, i.e., distinguishing left- and right-hand roles within an ongoing high-level action. Rather than modifying the decoder or LLM parts, our approach intervenes at the projection stage between the vision encoder and the language model. This is demonstrated in \Cref{fig:model_architecture}, where the video frames are encoded into spatiotemporal tokens, refined by \projectionmodule, and then projected into the LLM space for caption generation. Specifically, \projectionmodule conditions visual token re-weighting on the task description to amplify action-relevant spatial cues, while \newloss enforces structured supervision by masking auxiliary metadata and upweighting object-grounded tokens. Together, these components bridge the gap between generic visual representations and the fine-grained sub-action semantics required for accurate hand-role disambiguation.

\subsubsection{Text-Video features alignment.}

Sub-action detection requires preserving fine-grained spatial–temporal cues while aligning them with linguistic semantics describing hand roles. Standard linear or MLP projection layers \cite{Maaz2023VideoChatGPT, Maaz2024VideoGPT+} treat visual tokens uniformly, limiting their ability to emphasise sub-action–relevant evidence. To address this, we introduce \projectionmodule, a module that performs a more structured cross-modal interaction, with visual and textual information exchanged via attention mechanism and refined via a gated residual connection. Rather than compressing visual tokens into fixed queries \cite{li2023blip2, NEURIPS2022_960a172b} or propagating them unchanged \cite{Maaz2023VideoChatGPT, Maaz2024VideoGPT+}, \projectionmodule re-weights existing video tokens conditioned on textual context. Formally, let the video encoder produce a sequence of embeddings $\mathbf{V} \in \mathbb{R}^{B \times N_v \times d_v}$ and the language model provide token embeddings $\mathbf{T} \in \mathbb{R}^{B \times N_t \times d_t}$. The video features are first adapted to the text embedding dimension using a learnable projection, Elementary Video Adapter:


\begin{equation}
    \hat{\mathbf{V}} = \mathrm{MLP}(\mathrm{LN}(\mathbf{V})) \in \mathbb{R}^{B \times N_v \times d_t}
\end{equation}

To preserve positional consistency across frames, a learnable positional encoding $\mathbf{P}_v$ is added:

\begin{equation}
    \mathbf{V}' = \hat{\mathbf{V}} + \mathbf{P}_v.
\end{equation}

Next, cross-modal alignment is achieved through an attention mechanism that computes mappings between video and text sequences.
Specifically, the text-to-video attention context is defined as:

\begin{equation}
    \mathbf{A}_{t \rightarrow v} = \mathrm{softmax}\left( \frac{\mathbf{Q}_V \mathbf{K}_T^{\top}}{\sqrt{d_h}} + \mathbf{M}_t \right),
\end{equation}
where $\mathbf{Q}_V = \mathbf{W}_q \mathbf{V}'$, $\mathbf{K}_T = \mathbf{W}_k \mathbf{T}$, $d_h$ is the attention head dimension, 
and $\mathbf{M}_t$ is an additive attention mask for the text tokens. 

The attention output is then computed as:
\begin{equation}
    \mathbf{O}_V = \mathbf{A}_{t \rightarrow v} \mathbf{V}_T, \quad \text{where } \mathbf{V}_T = \mathbf{W}_v \mathbf{T}.
\end{equation}

The resulting video-attended representation is integrated via a gated residual connection:

\begin{equation}
    \tilde{\mathbf{V}} = \mathbf{V}' + \tanh(g_v) \cdot \mathbf{O}_V,
    \label{eq:res_gating}
\end{equation}
where $g_v$ is a learnable scalar gate that controls the fusion strength. This connection retains the original visual representation alongside the text-guided attention output, while the learnable gate prevents premature over-reliance on textual context early in training, allowing the model to gradually shift focus toward action-relevant visual cues.

Finally, a feed-forward refinement is applied to re-align the fused representation with the LLM’s token embedding statistics:
\begin{equation}
    \mathbf{Z}_V = \tilde{\mathbf{V}} + \mathrm{FFN}(\mathrm{LN}(\tilde{\mathbf{V}})).
\end{equation}

Where FFN corresponds to the re-alignment module in \Cref{fig:model_architecture}. The residual connection here ensures the cross-modal fusion is preserved while the FFN non-linearly re-projects the fused features into the LLM's expected token embedding space without overwriting the hand–object grounding signals accumulated in the previous step.

This task-conditioned alignment selectively amplifies visual evidence relevant to hand-specific sub-actions, improving role disambiguation without altering the decoder or token budget. \Cref{fig:model_architecture} shows how the \projectionmodule refines video embeddings via input text. Also, the forward pass of the \projectionmodule\ is detailed in the supplementary material.

\subsubsection{Selective Supervision for Sub-Action Learning}
\label{subsubsec:method_metaloss}

Conventional language modelling losses assign equal importance to all tokens, regardless of their semantic or structural role. However, our training data contains both natural-language captions and structured metadata spans (e.g., \texttt{<meta>}\ldots\texttt{</meta>}) that describe auxiliary information such as object positions or spatial annotations. These metadata tokens are not intended to contribute directly to the language modelling objective. Prior selective training strategies~\cite{pmlr-v162-mindermann22a_rho_loss, NEURIPS2024_3322a9a7rho1} address token or sample importance through adaptive weighting or prioritisation mechanisms, often relying on auxiliary scoring procedures or reference models. In contrast, we introduce \textbf{\newloss}, a lightweight supervision strategy that selectively masks metadata spans and upweights object-related tokens to emphasise visually grounded semantics. Importantly, our method requires no auxiliary reference model, as masking positions are determined directly from the annotation structure during dataset curation.

Let the model produce logits $\mathbf{y} \in \mathbb{R}^{B \times T \times V}$ and the corresponding target tokens be 
$\mathbf{t} \in \{0, \dots, V-1\}^{B \times T}$, where $B$ is the batch size, $T$ is the sequence length, and $V$ is the vocabulary size. 
We define a binary mask $m_i \in \{0, 1\}$ for each token position $i$ as:
\begin{equation}
m_i =
\begin{cases}
0, & \text{if } t_i \text{ lies in meta span},\\[3pt]
1, & \text{otherwise.}
\end{cases}
\end{equation}

This mask excludes tokens between \texttt{<meta>} and \texttt{</meta>} from loss computation, preventing the model from wasting capacity on structural markup and encouraging it to focus on semantically meaningful content (actions, hands, and objects). We construct the object set $O$ by taking the top-$K$ object nouns from BiScopeCap captions and mapping them to tokenizer IDs. This yields a compact vocabulary of frequently manipulated objects (e.g. pan, tap, knife, mug) rather than a large open-vocabulary lexicon. Next, we introduce a per-token weighting term $w_i$ to emphasise object mentions:
\begin{equation}
w_i =
\begin{cases}
\lambda_o, & \text{if } t_i \in \mathcal{O},\\[3pt]
1, & \text{otherwise,}
\end{cases}
\end{equation}
Where $\mathcal{O}$ denotes the set of vocabulary indices corresponding to object terms derived from metadata provided with the ground-truth captions, and $\lambda_o > 1$ is a coefficient controlling the degree of emphasis. We set $\lambda_o=3$ with ablation results comparing \newloss\ to standard cross-entropy shown in Table~\ref{tab:ablation_biscopenet_components}.
This weighting mechanism increases the gradient magnitude associated with visually grounded words, thereby strengthening multimodal alignment. Token-level cross-entropy for position $i$ is computed as:
\begin{equation}
\mathrm{CE}(\mathbf{y}_i, t_i)
= -\log \mathrm{softmax}(\mathbf{y}_i)_{t_i}.
\end{equation}
Combining masking and weighting, the overall \newloss\ objective is formulated as:
\begin{equation}
\mathcal{L}_{\text{meta}} =
\frac{\sum_i m_i w_i \, \mathrm{CE}(\mathbf{y}_i, t_i)}{\sum_i m_i w_i}.
\end{equation}

Intuitively, \newloss extends standard cross-entropy by (i) masking <meta> spans (e.g., structured punctuation or bounding-box markup) and (ii) upweighting object tokens to emphasise semantically meaningful content. It requires no modification to the optimiser or logits, relying only on per-token scalar weights and a binary mask. This encourages focus on visually relevant tokens while maintaining stable, normalised optimisation. Because \newloss reweights terms based on textual metadata rather than instance-specific patterns, it is data-agnostic and applicable to captioning tasks with structured targets. The complete algorithm is provided in the supplementary material.

\section{Experiments}
\label{sec:experiments}

We introduce \textbf{\datasetname} and evaluate our framework for sub-action detection in egocentric video captioning. We first outline model configurations and implementation details, followed by dataset description and experimental results. 

\noindent\textbf{Implementation Details:}
All experiments are implemented in PyTorch and HuggingFace Transformers and trained on three NVIDIA A100 GPUs. We evaluate \modelname\ across multiple VLM backbones, including NVILA~\cite{nvila2025} and VideoChatGPT~\cite{Maaz2023VideoChatGPT}. Unless otherwise stated, all models are trained using AdamW (without weight decay) with a learning rate of $2 \times 10^{-5}$ and batch size 8. The train--test split follows~\Cref{tab:dataset_full_comparison}. Also, the MetaLoss weighting coefficient $\lambda_o$ and evaluation prompts were selected using a validation split and were fixed before evaluation on the test set.

For NVILA, which employs Qwen2-VL~\cite{Qwen2VL} as the LLM and SigLIP~\cite{siglip} as the vision encoder, we fine-tune the model for 3 epochs using 8 uniformly sampled frames per video. For VideoChatGPT, which employs Vicuna 1.5 \cite{chiang2023vicuna} as the LLM and CLIP~\cite{clip} as the vision encoder, we train for 12 epochs using 100 uniformly sampled frames per video. \modelname\ is integrated into these architectures without modifying the core language or vision encoders. During fine-tuning, both the vision encoder and the language model remain frozen, and only \projectionmodule\ is trained, ensuring that performance gains arise from improved video-text alignment rather than backbone optimisation. Additionally, we conduct an ablation study on the effect of different vision encoders within the VideoChatGPT backbone, as detailed in \Cref{subsubsec:ablate_vision_encoder}.

\subsection{BiScopeCap dataset}
\label{subsec:dataset}
Action recognition benchmarks span egocentric and exocentric settings. Charades-Ego \cite{sigurdsson2018charadesegolargescaledatasetpaired} provides paired first- and third-person videos for viewpoint transfer but focuses on high-level action categories without modelling hand-specific roles. Ego-Exo4D \cite{egoexo4d} further supports action recognition and human–object interaction across paired views. Within the egocentric domain, the EPIC-KITCHENS family \cite{epic_Damen2018EPICKITCHENS, epic_Damen2021PAMI, epic_Damen2022RESCALING, epic_VISOR2022} provides large-scale recordings of unscripted daily activities and has become a standard benchmark for action recognition and anticipation. Similarly, HD-Epic \cite{perrett2025hdepic} extends these annotations with richer multimodal signals and detailed descriptions. While these datasets enable high-level action recognition, they are not designed for sub-action detection, which requires explicit left–right hand role modelling. As shown in~\Cref{tab:dataset_full_comparison}, they lack hand-role attribution and spatially grounded video-level captions capturing fine-grained manipulation semantics. The closest effort, HOI-QA, provides frame-level hand–object annotations for selected frames only, without temporally coherent video-level supervision. To address this gap, we introduce \datasetname, which provides dense video-level captions that describe high-level actions and each hand’s role, augmented with textual spatial metadata.

\noindent\textbf{Dataset Curation:}
\datasetname\ is built through a unified annotation pipeline to support sub-action detection. Instead of new recordings, we standardise and enrich data from EPIC-KITCHENS, EPIC-VISOR \cite{epic_Damen2018EPICKITCHENS, epic_Damen2022RESCALING, epic_Damen2021PAMI, epic_VISOR2022}, Ego4D \cite{ego4d}, and HD-Epic \cite{perrett2025hdepic}. Frame-level hand–object annotations from HOI-QA \cite{hoiqa_bansal2024hoi} are consolidated into temporally coherent, video-level captions using Gemini \cite{Google2025} only for structured summarisation. All captions are reformulated into a unified format integrating (i) high-level actions, (ii) fine-grained hand roles, and (iii) structured spatial metadata. This harmonised strategy produces a single, task-aligned corpus explicitly designed for sub-action detection.

To illustrate how our annotation strategy supports sub-action detection, we summarise each source’s contribution. EPIC-KITCHENS and Ego4D provide narrations and temporally localised actions, while EPIC-VISOR adds segmentation masks that HOI-QA uses to extract bounding boxes for hands and objects along with keyframe captions. These signals, narrations, temporal boundaries, and hand–object regions, are unified into structured video-level captions using Google Gemini 2.0-Flash \cite{Google2025}, conditioned on human narrations and annotated regions to preserve explicit grounding and accurate hand-role attribution. For HD-Epic, although it includes hand motion descriptions in video-level, these often omit the high-level action and lack structured metadata. We therefore reformulate and augment its annotations using available hand and object segmentations to align with our unified sub-action–oriented caption format.

\begin{table*}[t]
\centering
\small
\setlength{\tabcolsep}{6pt}
\renewcommand{\arraystretch}{1.1}
\caption{Comparison between existing egocentric datasets and our proposed \textbf{\datasetname}, along with the distribution of samples across its constituent subsets.}
\label{tab:dataset_full_comparison}
\resizebox{\textwidth}{!}{
\begin{tabular}{lcccccc|cc}
\toprule
\textbf{Dataset} & \textbf{Action} & \textbf{Hand Role} & \textbf{Spatial} & \textbf{Coarse-grained} & \textbf{Fine-grained} & \textbf{Domain} & \textbf{\# Train} & \textbf{\# Test} \\
 & \textbf{Label} & \textbf{Labels} & \textbf{Metadata} & \textbf{Captions} & \textbf{Captions} & \textbf{Source} & \textbf{Samples} & \textbf{Samples} \\
\midrule
Epic-Kitchens / Visor & \textcolor{ForestGreen}{\textcolor{ForestGreen}{\cmark}} & \textcolor{BrickRed}{\xmark} & \textcolor{BrickRed}{\xmark} & \textcolor{BrickRed}{\xmark} & \textcolor{BrickRed}{\xmark} & Kitchen / Daily Tasks & 31,432 & 3,729 \\
Ego4D & \textcolor{ForestGreen}{\cmark} & \textcolor{BrickRed}{\xmark} & \textcolor{BrickRed}{\xmark} & \textcolor{BrickRed}{\xmark} & \textcolor{BrickRed}{\xmark} & Diverse Activities & 22,916 & 3,304 \\
HOI-QA & \textcolor{ForestGreen}{\cmark}  & \textcolor{BrickRed}{\xmark} & \textcolor{ForestGreen}{\cmark} & \textcolor{BrickRed}{\xmark} & \textcolor{ForestGreen}{\cmark} & Kitchen / Daily Tasks & – & – \\
HD-Epic & \textcolor{ForestGreen}{\cmark} & \textcolor{ForestGreen}{\cmark} & \textcolor{BrickRed}{\xmark} & \textcolor{BrickRed}{\xmark} & \textcolor{ForestGreen}{\cmark} & Kitchen / Daily Tasks & 38,520 & 2,846 \\
\cdashline{1-9}
\textbf{\datasetname (ours)} & \textcolor{ForestGreen}{\cmark} & \textcolor{ForestGreen}{\cmark} & \textcolor{ForestGreen}{\cmark} & \textcolor{ForestGreen}{\cmark} & \textcolor{ForestGreen}{\cmark} & Diverse Activities & \textbf{92,868} & \textbf{9,879} \\
\bottomrule
\end{tabular}}
\end{table*}


Clips are segmented by narration timestamps and resized to the model’s input resolution, with bounding boxes temporally and spatially re-aligned to preserve accurate grounding. \datasetname\ is constructed solely from research-licensed datasets (EPIC-KITCHENS, Ego4D, HD-Epic, HOI-QA). To address concerns about annotation hallucination, we conducted a manual audit of 200 randomly sampled training clips balanced across datasets. We observed that $\approx 5\%$ contain minor descriptive inaccuracies (e.g. imprecise verbs), while $\approx 8\%$ and $\approx 10\%$ exhibit left and right hand hallucination, respectively. We consider this level of noise acceptable, as the model is intended to learn generalised grounding patterns rather than memorise individual annotations. Similar annotation noise has been reported in other large-scale egocentric datasets~\cite{ego4d}, where large-scale manual or semi-automatic labelling inevitably introduces errors and biases. All derived annotations and curation scripts will be released, while the raw videos remain under their original licences.

\begin{table}[t]
\centering
\caption{Comparison between LLM-as-a-judge scores and human evaluation over 200 predicted captions.}
\label{tab:manual_scoring_comparison}
\begin{tabular}{lccccc}
\toprule
Method & \# Samples & \% ActRec $\uparrow$ & \% RH $\uparrow$ & \% LH $\uparrow$ & \% ObjRec $\uparrow$ \\
\midrule
GPT Scoring   & 200 & 44.5 & 31.0 & 36.0 & 47.0 \\
Human Scoring & 200 & 42.0  & 30.3 & 35.6 & 51.0 \\
\bottomrule
\end{tabular}
\end{table}

\noindent\textbf{Dataset Statistics:}

We summarise key statistics highlighting \datasetname’s suitability for sub-action detection (see~\Cref{tab:dataset_full_comparison}). It contains 11,371 unique action classes (verb–noun pairs) and 5,338 active objects (including hands), covering diverse manipulation scenarios. Analysing left/right-hand and bimanual interactions confirms balanced supervision for hand-role disambiguation. Object frequency further demonstrates broad coverage across contexts. These statistics show that \datasetname provides diverse, grounded supervision for fine-grained sub-action understanding beyond high-level action classification. Additional details are in the supplementary.

\subsection{Metric Scoring System}
\label{sec:score}

To evaluate performance on sub-action detection, we assess model outputs along four complementary dimensions: (1) high-level action recognition (\textbf{ActRec}), (2) right-hand role recognition (\textbf{RH}), (3) left-hand role recognition (\textbf{LH}), and (4) object recognition (\textbf{ObjRec}). Together, these metrics assess whether a model correctly and explicitly captures the high-level action, along with the left- and right-hand roles, and their interactions with objects that define sub-actions. Accuracy is used for all dimensions.

We adopt an LLM-as-a-judge evaluation protocol using GPT-5 Mini \cite{OpenAI2025} to score generated captions across these categories. Importantly, caption evaluation relies on a separate model from \datasetname generation, ensuring independence between data construction and assessment. The full evaluation prompt is provided in the supplementary material. To verify reliability, we manually scored 200 randomly sampled predicted captions from all subsets and compared these with human judgments and GPT-5-based scores. The automatic evaluation closely matches human assessment, with overall accuracy differing by less than one percentage point and hand- and object-specific metrics varying by only 2–3 points on average (see \Cref{tab:manual_scoring_comparison}). These results indicate that the LLM judge provides a consistent and reliable assessment of the correctness of fine-grained sub-actions.

Beyond agreement with human judgments, we further assess the stability of our evaluation protocol itself. We repeat the full pipeline, both model inference and LLM-based scoring, three times for our best-performing configuration (NVILA backbone) and report the mean and standard deviation across runs in Table~\ref{tab:stability}. The low variance across all four metrics indicates that both BiScopeNet's predictions and the GPT-5 Mini scoring procedure are stable, and that the gains reported in \Cref{tab:performance_biscope_subset} are not an artefact of evaluation noise.

\begin{table}[t]
\centering
\small
\caption{\modelname stability results across three independent inference and scoring runs on NVILA backbone.}
\label{tab:stability}
\begin{tabular}{lcccc}
\toprule
Method & \% ActRec $\uparrow$ & \% RH $\uparrow$ & \% LH $\uparrow$ & \% ObjRec $\uparrow$ \\
\midrule
BiScopeNet (avg.\ 3 runs) & 45.3 & 44.7 & 40.6 & 58.9 \\
Standard Deviation        & $\pm$0.2 & $\pm$0.5 & $\pm$0.6 & $\pm$0.07 \\
\bottomrule
\end{tabular}
\end{table}

\Cref{tab:manual_scoring_comparison} shows that the GPT-based evaluation closely aligns with human judgements across all considered metrics. The differences are minimal, typically within a few percentage points, confirming that automated scoring reliably reflects fine-grained performance in hand-role and object recognition. As sub-action detection requires structurally coherent captions that explicitly separate high-level actions and hand-specific roles, invalid or unstructured outputs can directly compromise evaluation. Vision–language models occasionally generate captions that are grammatically inconsistent, improperly formatted, or contain extraneous tags (see \Cref{fig:qualitative_result_caption_structure1}). We therefore apply an additional filtering prompt using GPT-5 Mini \cite{OpenAI2025} (provided in the supplementary material) to identify such outputs. The frequency of these cases is recorded as an indicator of model stability under structured sub-action supervision. These unstructured captions are treated as incorrect predictions, as they fail to provide evaluable hand-role or object descriptions. For completeness, we also report scores without penalising such samples, illustrating how structured-caption compliance influences overall sub-action detection performance (see the supplementary material).

\begin{table*}[t]
\centering
\small
\setlength{\tabcolsep}{5pt}
\caption{Comparison of \textbf{\modelname} with state-of-the-art VLMs on BiScopeCap. \textbf{\textcolor{blue}{Blue}} denote the best results.}
\label{tab:performance_biscope_subset}
\resizebox{\textwidth}{!}{%
\begin{tabular}{l|c|c|c|cccc|c}
\toprule
\textbf{Method} & \textbf{Vision Encoder} & \textbf{Backbone} & \textbf{Trained} & \textbf{\% ActRec {$\uparrow$}} & \textbf{\% RH {$\uparrow$}} & \textbf{\% LH {$\uparrow$}} & \textbf{\% ObjRec {$\uparrow$}} & \textbf{\# Unstruct {$\downarrow$}} \\
\midrule
SmolVLM \cite{marafioti2025smolvlm} & SigLIP \cite{siglip} & - & \xmark & 10.4 & 11.8 & 14.6 & 27.6 & 4505 \\
Qwen 2.5-VL \cite{bai2025qwen25vl} & - & - & \xmark & 25.6 & 29.9 & 37.0 & 54.0 & 1 \\
Qwen 3.5 \cite{qwen2025qwen35} & - & - & \xmark & 26.0 & 32.2 & 40.6 & 53.3 & 6 \\

Intern VL 2.5 \cite{chen2024expanding} & InternViT \cite{chen2024internvl} & - & \xmark & 24.8 & 27.4 & 37.0 & 51.5 & 5 \\
NVILA \cite{nvila2025} & SigLIP & - & \xmark & 14.1 & 17.2 & 28.7 & 39.7 & 0 \\
NVILA \cite{nvila2025} & SigLIP & - & \cmark & 44.0 & 43.0 & 39.1 & 58.0 & 0 \\
Qwen 3.5 \cite{qwen2025qwen35} & - & - & \cmark & 41.8 & 42.0 & 40.0 & 54.3 & 40 \\
VideoChatGPT \cite{Maaz2023VideoChatGPT} & CLIP \cite{clip} & - & \cmark & 14.3 & 13.7 & 17.7 & 30.1 & 488 \\
\cdashline{1-9}
\textbf{\modelname (Ours)} & CLIP & VideoChatGPT & \cmark & 30.7 & 19.4 & 29.6 & 47.9 & 7 \\
\textbf{\modelname (Ours)} & DINOv3 \cite{simeoni2025dinov3} & VideoChatGPT & \cmark & 33.4 & 31.9 & 31.4 & 53.5 & 5 \\
\textbf{\modelname (Ours)} & SigLIP & NVILA & \cmark & 
\textbf{\textcolor{blue}{45.5}} & 
\textbf{\textcolor{blue}{44.9}} & 
\textbf{\textcolor{blue}{41.2}} & 
\textbf{\textcolor{blue}{58.9}} & 
0 \\
\bottomrule
\end{tabular}%
}
\end{table*}

\subsection{BiScopeNet Performance Analysis}
\label{subsec:biscopenet_perf_analysis}
For fair comparison, we evaluate across both zero-shot and supervised configurations. We compare against off-the-shelf VLMs, SmolVLM~\cite{marafioti2025smolvlm}, Qwen~\cite{qwen2025qwen35, bai2025qwen25vl}, InternVL~\cite{chen2024internvl}, and NVILA~\cite{nvila2025}, in zero-shot mode without task-specific adaptation on \datasetname. For Video-ChatGPT, zero-shot performance was consistently poor across all metrics, so we exclude this configuration from the main comparison. We further include supervised baselines, where models such as NVILA~\cite{nvila2025}, Qwen 3.5~\cite{qwen2025qwen35}, and VideoChatGPT~\cite{Maaz2023VideoChatGPT} are fine-tuned on \datasetname\ using their original architectures without integrating \modelname. Finally, we evaluate \modelname\ integrated into both VideoChatGPT and NVILA backbones, where the proposed modules are introduced while keeping the core language and vision backbones frozen and optimising only the projection-related components. This evaluation protocol allows us to distinguish improvements arising from general zero-shot capability, supervised task adaptation, and the additional gains introduced by \modelname.

\Cref{tab:performance_biscope_subset} reports results on the full \datasetname, which combines the Ego4D, Epic, and HD-Epic subsets for comprehensive evaluation across diverse egocentric settings. Overall, \modelname\ consistently outperforms competing VLM baselines across all evaluation metrics, including action recognition (ActRec), left/right hand understanding (LH, RH), object recognition (ObjRec), and structured caption generation (Unstruct).

Among the baseline architectures, the supervised version of NVILA~\cite{nvila2025} achieves a substantial improvement over its zero-shot counterpart, increasing ActRec from 14.1\% to 44.0\%, which highlights the effectiveness of \datasetname\ for fine-grained action and sub-action learning. Integrating \modelname\ into the NVILA backbone yields further gains across all metrics, achieving the strongest overall performance among evaluated configurations, with no malformed generations observed on this benchmark. These improvements demonstrate that the proposed \projectionmodule\ and \newloss\ provide complementary benefits beyond standard supervised adaptation and generalise effectively across different VLM architectures.

To further evaluate backbone generalisability, we additionally integrate \modelname\ into VideoChatGPT~\cite{Maaz2023VideoChatGPT}. Using the same CLIP vision encoder as the original VideoChatGPT model, \modelname\ improves ActRec from 14.3\% to 30.7\% and ObjRec from 30.1\% to 47.9\%, while simultaneously reducing unstructured outputs from 488 to only 7. Replacing CLIP with DINOv3 further improves performance. These results indicate that \modelname\ not only improves reasoning and grounding within existing VLM backbones, but also benefits from stronger visual representations provided by more advanced vision encoders.

Also, we considered the efficiency analysis of FocusAlign. Compared to standard linear and MLP projectors, it adds fewer than 0.3B parameters and less than 1\% additional TFLOPs, while having negligible impact on GPU memory usage and inference latency. \Cref{tab:efficiency} indicates that the performance improvements achieved by FocusAlign come with virtually no meaningful efficiency cost.

\begin{table}[t]
\centering
\footnotesize
\caption{Efficiency comparison of projection variants, showing minimal FocusAlign overhead.}
\label{tab:efficiency}
\begin{tabular}{llcccc}
\toprule
Model & Proj. & Params (B) & TFLOPs & GPU (GB) & Time (s) \\
\midrule
VCG   & Linear     & 7.05 & 45.44 & 14.68 & 3.20 \\
VCG   & MLP        & 7.06 & 45.45 & 14.71 & 3.12 \\
VCG   & FocusAlign & 7.30 & 45.71 & 15.17 & 3.12 \\
NVILA & MLP        & 8.09 & 38.53 & 15.78 & 3.89 \\
NVILA & FocusAlign & 8.23 & 38.86 & 16.03 & 3.95 \\
\bottomrule
\end{tabular}
\end{table}

\subsection{Ablation Studies}

\begin{table*}[!t]
\centering
\footnotesize
\setlength{\tabcolsep}{3pt}
\renewcommand{\arraystretch}{1.1}
\caption{Ablation study of \textbf{\modelname} components using NVILA backbone \cite{nvila2025} on \datasetname. The proposed \projectionmodule\ and \newloss\ consistently improve all metrics, demonstrating complementary benefits for sub-action understanding. \textbf{\textcolor{blue}{Blue}} indicates the best results.}
\label{tab:nvila_ablation_biscopenet_components}
\resizebox{\textwidth}{!}{%
\begin{tabular}{l|c|c|c|cccc|c}
\toprule
\textbf{Method} & \textbf{Component} & \textbf{Vision Encoder} & \textbf{Backbone} & \textbf{\% ActRec \textsuperscript{$\uparrow$}} & \textbf{\% RH \textsuperscript{$\uparrow$}} & \textbf{\% LH \textsuperscript{$\uparrow$}} & \textbf{\% ObjRec \textsuperscript{$\uparrow$}} & \textbf{\# Unstruct \textsuperscript{$\downarrow$}} \\
\midrule
NVILA & MLP + Cross Entropy & SigLIP & - & 44.0	& 43.0 & 39.1 & 58.0 & 0 \\
\modelname & MLP + \newloss & SigLIP & NVILA & 44.0 & 44.1 & 39.1	& 58.7 & 0 \\
\modelname & \projectionmodule~ + \newloss~ & SigLIP & NVILA & \textbf{\textcolor{blue}{45.5}}	& \textbf{\textcolor{blue}{44.9}} & \textbf{\textcolor{blue}{41.2}} & \textbf{\textcolor{blue}{58.9}} & 0 \\

\bottomrule
\end{tabular}}
\end{table*}

\begin{table*}[!t]
\centering
\footnotesize
\setlength{\tabcolsep}{3pt}
\renewcommand{\arraystretch}{1.1}
\caption{Ablation study of \textbf{\modelname} components using VideoChatGPT backbone \cite{Maaz2023VideoChatGPT} on the Epic subset of \datasetname. The proposed \projectionmodule\ and \newloss\ consistently deliver the best performance, highlighting their complementary benefits for sub-action understanding. \textbf{\textcolor{blue}{Blue}} indicates the best results.}
\label{tab:ablation_biscopenet_components}
\resizebox{\textwidth}{!}{%
\begin{tabular}{l|c|c|cccc|c}
\toprule
\textbf{Method} & \textbf{Component} & \textbf{Vision Encoder} & \textbf{\% ActRec \textsuperscript{$\uparrow$}} & \textbf{\% RH \textsuperscript{$\uparrow$}} & \textbf{\% LH \textsuperscript{$\uparrow$}} & \textbf{\% ObjRec \textsuperscript{$\uparrow$}} & \textbf{\# Unstruct \textsuperscript{$\downarrow$}} \\
\midrule
VideoChatGPT & Linear + Cross Entropy & CLIP & 12.8 & 10.4 & 5.9 & 27.3 & 374 \\
VideoChatGPT & MLP + Cross Entropy & CLIP & 16.8 & 15.5 & 12.8 & 26.4 & 437 \\
\modelname & \projectionmodule~ + Cross Entropy & CLIP & 13.9 & 10.6 & 12.9 & 18.8 & 196 \\
\modelname & Linear + \newloss & CLIP &  19.6 & 19.0 & 25.8 & 40.4 & 0 \\
\modelname & \projectionmodule~ + \newloss~ & CLIP & \textbf{\textcolor{blue}{32.5}} & 18.4 & 25.8 & 46.4 & 5 \\
\modelname & \projectionmodule~ +  \newloss~ + CLS & CLIP & 30.7 & \textbf{\textcolor{blue}{19.4}} & \textbf{\textcolor{blue}{29.6}} & \textbf{\textcolor{blue}{47.9}} & 7 \\

\bottomrule
\end{tabular}}
\end{table*}

\subsubsection{Model Architectures.}
\label{subsubsec:model_arch}

\Cref{tab:nvila_ablation_biscopenet_components} presents an ablation study of \modelname\ components on \datasetname\ using NVILA as the backbone. Applying \newloss\ alone with the standard MLP projection already yields modest improvements over the original NVILA configuration, particularly in RH and ObjRec performance. Incorporating the proposed \projectionmodule\ together with \newloss\ produces the strongest and most consistent gains across all evaluation metrics. These results demonstrate the complementary roles of \projectionmodule\ and \newloss\ in improving fine-grained sub-action understanding and visually grounded reasoning.
To investigate if our approach generalises across different baselines, we tested our approach on the weakest baseline, VideoChatGPT \cite{Maaz2023VideoChatGPT}, on the Epic subset of \datasetname; the results are reported in \Cref{tab:ablation_biscopenet_components}. Replacing the linear projection in VideoChatGPT \cite{Maaz2023VideoChatGPT} with an MLP improves the overall accuracy, indicating that a deeper mapping better aligns modalities. However, using \projectionmodule\ alone yields inconsistent gains, improving hand-related metrics but reducing object accuracy, suggesting that \projectionmodule\ without proper guidance may overfocus on local cues. When coupled with the proposed \newloss, performance increases substantially across all dimensions, while the unstructured error drops sharply, confirming a strong correlation between feature selection and loss supervision. Adding a CLS token slightly stabilises hand and object recognition, but marginally lowers global accuracy, implying a trade-off between summarisation and fine-grained diversity. Overall, the \projectionmodule–\newloss\ combination proves essential for robust video–text alignment and neither component alone yields the same improvement.

We additionally isolate the contribution of \newloss against alternative losses, namely Label Smoothing and Focal Loss, under the same \projectionmodule configuration, as well as the effect of the residual gating mechanism within \projectionmodule. Both studies consistently support our design choices; the corresponding results are provided in the supplementary material.

\begin{table*}[ht]
\centering
\small
\setlength{\tabcolsep}{4pt}
\renewcommand{\arraystretch}{1.1}
\caption{Comparison of \textbf{\modelname} using VideoChatGPT~\cite{Maaz2023VideoChatGPT} as backbone with state-of-the-art VLMs across \textbf{Ego4D}, \textbf{Epic+Ego4D}, and \textbf{HD-Epic} subsets of \datasetname. \# Unstruct counts are over the corresponding subset test split, not the full \datasetname test set. \textbf{\textcolor{blue}{Blue}} denote the best results.}
\label{tab:performance_other_subsets}
\resizebox{\textwidth}{!}{%
\begin{tabular}{ll|c|c|cccc|c}
\toprule
&\textbf{Method}  & \textbf{Trained} & \textbf{Vision Encoder} & \textbf{\% ActRec\textsuperscript{$\uparrow$}} & \textbf{\% RH\textsuperscript{$\uparrow$}} & \textbf{\% LH \textsuperscript{$\uparrow$}} & \textbf{\% ObjRec\textsuperscript{$\uparrow$}} & \textbf{\# Unstruct\textsuperscript{$\downarrow$}} \\
\midrule

\multirow{4}{*}{\rotatebox{90}{\makebox[0pt][c]{\textbf{Ego4D}}}}
&SmolVLM \cite{marafioti2025smolvlm}  & \xmark & SigLIP  & 8.9  & 12.0 & 16.5 & 25.8 & 1256 \\
&NVILA \cite{nvila2025} & \xmark & SigLIP  & 13.0 & 16.7 & 30.9 & 39.7 & 0 \\
&VideoChatGPT \cite{Maaz2023VideoChatGPT}  & \cmark & CLIP  & 7.2   & 10.1 & 12.3 & 24.0 & 563 \\
\cdashline{2-9}
&\textbf{\modelname (Ours)}  
& \cmark & DINOv3  & \textbf{\textcolor{blue}{22.1}} & \textbf{\textcolor{blue}{38.7}} & \textbf{\textcolor{blue}{39.6}} & \textbf{\textcolor{blue}{42.1}} & 46 \\
\midrule

\multirow{4}{*}{\rotatebox{90}{\makebox[0pt][c]{\textbf{Epic+Ego4D}}}}
&SmolVLM \cite{marafioti2025smolvlm} & \xmark & SigLIP  & 13.1 & 15.6 & 18.9 & 29.6 & 2785 \\
& NVILA \cite{nvila2025}  & \xmark & SigLIP  & 17.7 & 21.5 & 34.9 & 43.1 & 0 \\
& VideoChatGPT \cite{Maaz2023VideoChatGPT} & \cmark & CLIP   & 14.5 & 20.7 & 22.6 & 31.0 & 380 \\
\cdashline{2-9}
&\textbf{\modelname (Ours)}    & \cmark & DINOv3 & \textbf{\textcolor{blue}{33.5}} & \textbf{\textcolor{blue}{33.4}} & \textbf{\textcolor{blue}{34.9}} & \textbf{\textcolor{blue}{53.6}} & 4 \\
\midrule

\multirow{4}{*}{\rotatebox{90}{\makebox[0pt][c]{\textbf{HD-Epic}}}}
&SmolVLM \cite{marafioti2025smolvlm}  & \xmark & SigLIP  & 6.4  & 5.9  & 8.2  & 24.6 & 1720 \\
&NVILA \cite{nvila2025}  & \xmark & SigLIP  & 8.8  & 10.7 & \textbf{\textcolor{blue}{19.3}} & 34.5 & 0 \\
&VideoChatGPT \cite{Maaz2023VideoChatGPT}       & \cmark & CLIP   & 7.1  & 8.2 & 8.5 & 28.8 & 0 \\
\cdashline{2-9}
&\textbf{\modelname (Ours)} & \xmark & DINOv3  & 10.1 & 10.7 & 15.8 & 24.0 & 0 \\
&\textbf{\modelname (Ours)}    & \cmark & DINOv3  & \textbf{\textcolor{blue}{28.5}} & \textbf{\textcolor{blue}{21.4}} & 15.3 & \textbf{\textcolor{blue}{57.2}} & 10 \\
\bottomrule
\end{tabular}%
}
\end{table*}

\subsubsection{Ablation on Dataset Subsets.}
\label{subsubsec:ablation_ds_subsets}
To analyse the robustness of sub-action detection under different supervision sources, we train and evaluate \modelname with the VideoChatGPT backbone separately on each subset of \datasetname: Ego4D, Epic+Ego4D, and HD-Epic (see~\Cref{tab:performance_other_subsets}). Across all splits, \modelname\ consistently achieves the strongest performance in nearly every metric. On the Ego4D and Epic+Ego4D subsets, the model demonstrates particularly strong left- and right-hand recognition accuracy, indicating effective learning of fine-grained hand-role distinctions. On HD-Epic, improvements are especially pronounced in object recognition, reflecting reliable grounding in visually dense and high-resolution scenarios. The consistency of these gains across subsets with different characteristics suggests that the improvements stem from task-aligned cross-modal alignment and selective supervision, rather than overfitting to a particular data source. Overall, the results confirm that \modelname\ generalises sub-action detection across varied egocentric environments and interaction contexts. We also study the cross-subset generalisation capability of our approach by training \modelname on Epic+Ego4D and evaluating it on the HD-Epic subset. Despite not being trained on the HD-Epic distribution, our model achieves better results than the original VideoChatGPT, even when the latter is trained on HD-Epic. This demonstrates the generalisation ability of the proposed approach.

\begin{table*}[ht]
\centering
\footnotesize
\setlength{\tabcolsep}{3pt}
\renewcommand{\arraystretch}{1.1}
\caption{Ablation study of \textbf{BiScopeNet} integrated into the VideoChatGPT \cite{Maaz2023VideoChatGPT} backbone with different vision encoders (CLIP, DINOv3, and SigLIP2) and the presence or absence of the CLS token on the Epic subset of \datasetname. \modelname\ consistently improves performance across all encoders. SigLIP2+CLS achieves the highest overall and object-level accuracy, while DINOv3 shows superior sensitivity to hand-specific cues, demonstrating the adaptability of \modelname\ to diverse vision backbones. \textbf{\textcolor{blue}{Blue}} highlights the best results.}
\label{tab:ablation_vision_encoder_cls_comparison}
\resizebox{\textwidth}{!}{%
\begin{tabular}{l|c|c|cccc|c}
\toprule
\textbf{Method} & \textbf{Vision Encoder} & \textbf{CLS} & \textbf{\% ActRec \textsuperscript{$\uparrow$}} & \textbf{\% RH \textsuperscript{$\uparrow$}} & \textbf{\% LH \textsuperscript{$\uparrow$}} & \textbf{\% ObjRec \textsuperscript{$\uparrow$}} & \textbf{\# Unstruct \textsuperscript{$\downarrow$}}  \\
\midrule
VideoChatGPT & CLIP & \xmark & 12.8 & 10.4 & 5.9 & 27.3 & 374 \\
\modelname & CLIP & \xmark & 32.5 & 18.4 & 25.8 & 46.4 & 5 \\
\modelname & CLIP & \cmark & 30.7 & 19.4 & 29.6 & 47.9 & 7 \\
\modelname & DINOv3 & \xmark & 33.1 & 17.6 & 36.6 & 48.0 & 5 \\
\modelname & DINOv3 & \cmark & 28.2 & \textbf{\textcolor{blue}{23.0}} & \textbf{\textcolor{blue}{38.2}} & 49.4 & 13 \\
\modelname & SigLIP2 & \xmark & 32.4 & 21.6 & 31.8 & 47.0 & 5 \\
\modelname & SigLIP2 & \cmark & \textbf{\textcolor{blue}{38.1}} & 21.9 & 15.5 & \textbf{\textcolor{blue}{51.4}} & 3 \\
\bottomrule
\end{tabular}}
\end{table*}

\subsubsection{Ablation on Vision Encoder and CLS Token}
\label{subsubsec:ablate_vision_encoder}
We further conduct an ablation study by evaluating \modelname\ using VideoChatGPT as backbone with different vision encoders and with/without the CLS token to analyse the effect of visual feature representations on sub-action detection performance. Results on the Epic subset of \datasetname\ are reported in \Cref{tab:ablation_vision_encoder_cls_comparison}. We consider three widely used encoders, CLIP \cite{clip}, SigLIP2 \cite{tschannen2025siglip2multilingualvisionlanguage}, and DINOv3 \cite{simeoni2025dinov3}, and compare against the baseline VideoChatGPT \cite{Maaz2023VideoChatGPT} under identical training settings. Across all encoders, \modelname\ consistently outperforms the baseline, indicating that the gains arise from our task-conditioned alignment and selective supervision rather than vision-encoder-specific properties. While SigLIP2 with a CLS token achieves the strongest overall and object-level accuracies, DINOv3 shows competitive performance on hand-specific metrics, suggesting complementary strengths in semantic grounding and sensitivity to local motion.

The effect of the CLS token varies across encoders, modestly influencing high-level versus fine-grained metrics, but the relative improvement of \modelname\ remains stable. Overall, these results demonstrate that \modelname\ functions as a backbone- and vision-encoder-agnostic framework for sub-action detection, improving hand-role and object grounding without relying on a particular vision encoder.

\begin{figure*}[!t] 
\centering
\includegraphics[width=\linewidth]{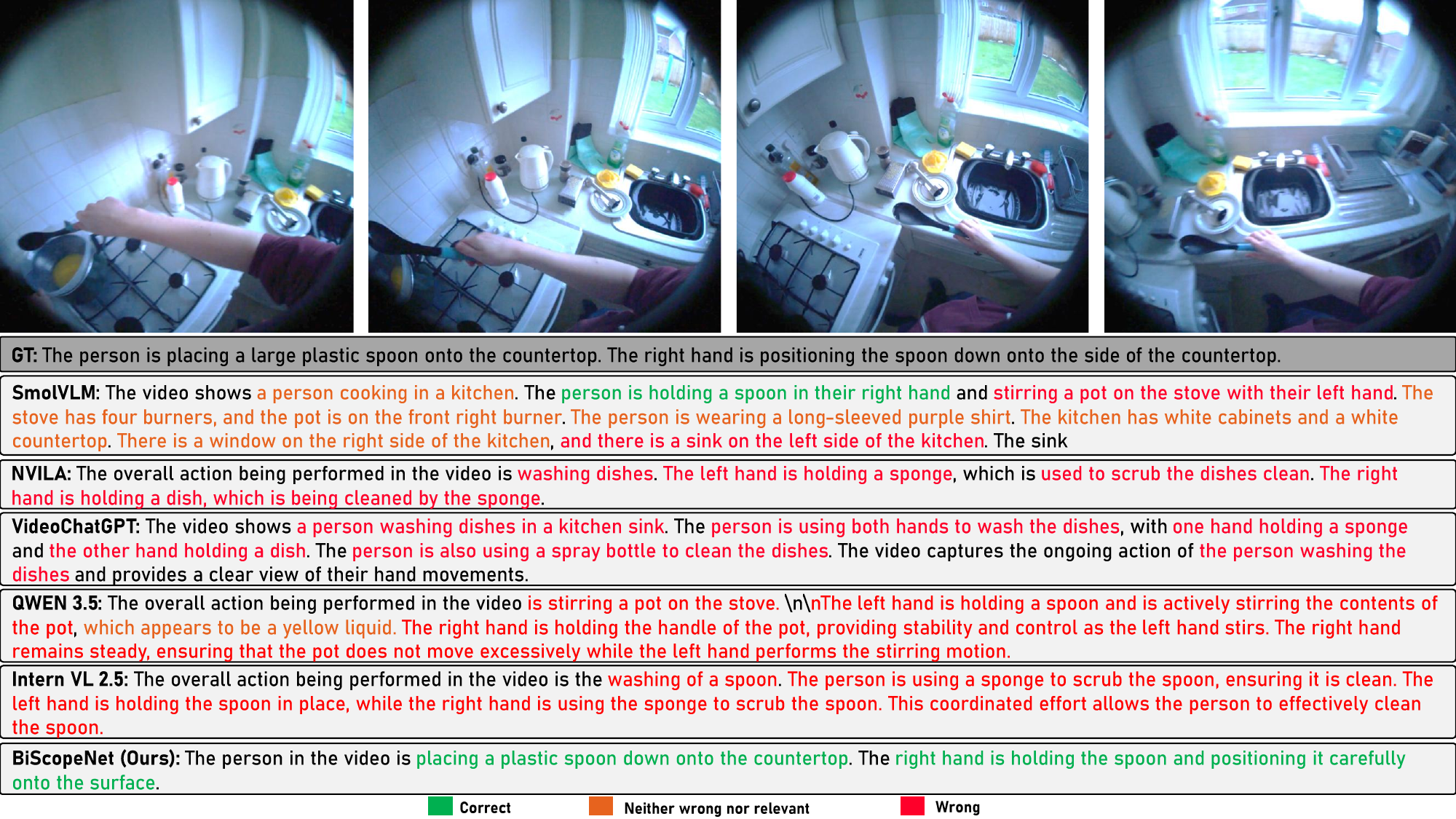}
\caption{Qualitative comparison of visual understanding across methods.}
\label{fig:qualitative_result_visual_understanding1}
\end{figure*}

\begin{figure*}[!t]
\centering
\includegraphics[width=0.85\linewidth]{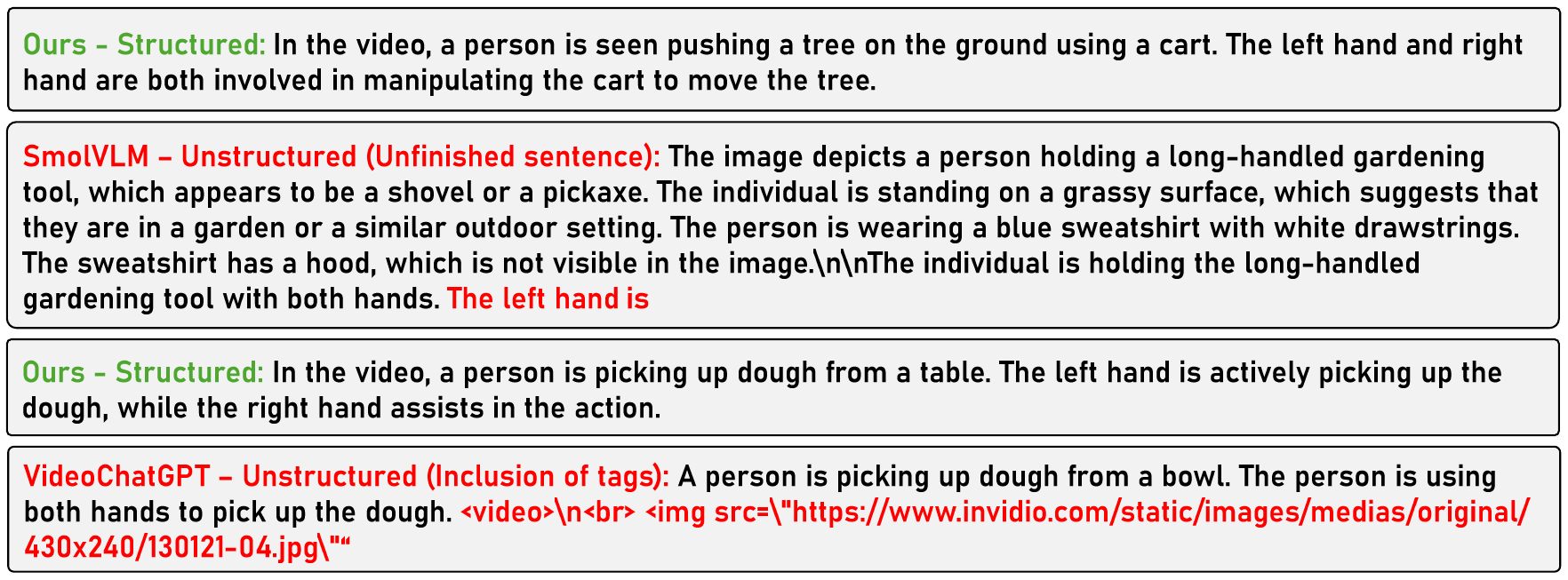}
\caption{Qualitative comparison of output structure across methods.}
\label{fig:qualitative_result_caption_structure1}
\end{figure*}

\subsection{Qualitative Results}

We present qualitative comparisons to highlight differences in fine-grained visual grounding and caption structure. As shown in \Cref{fig:qualitative_result_visual_understanding1}, baseline models often fail to correctly localise hands, distinguish their roles, or associate them with manipulated objects. In contrast, our approach (on VideoChatGPT backbone) produces more coherent hand–object–action reasoning, leading to semantically grounded sub-action descriptions. Additional examples are provided in the supplementary material. We also illustrate unstructured captioning failures in \Cref{fig:qualitative_result_caption_structure1}. Baseline outputs frequently contain incomplete sentences or extraneous artefacts (e.g., tags or URLs), which reduce clarity and structural consistency.

\section{Conclusion}
\label{sec:conclusion}

We presented \modelname, a framework for sub-action recognition in egocentric video captioning that models left– and right-hand roles through a selective text-guided projection module, \projectionmodule, coupled with an object-aware loss, \newloss. Together with the proposed dataset, \datasetname, which provides structured video-level supervision for high-level actions, hand roles, and spatial metadata, our approach promotes more consistent and grounded hand–, object–, and action- reasoning. Across multiple video descriptors, we observe consistent improvements in hand-specific accuracy, action recognition, and caption structure, further evaluated using an LLM-based protocol independently validated against 200 human-scored predictions.

\textbf{Limitations and future work.} \modelname\ samples a fixed number of frames per clip, which limits its ability to model sub-actions in long-term videos where hand roles evolve over extended temporal spans. We additionally observe two recurring failure modes (examples in the supplementary material): (i) repetitive, low-amplitude hand motions (e.g.\ successive up/down movements) introduce ambiguity that the model struggles to disambiguate, and (ii) very short clips combined with high camera motion sharply degrade caption quality, a limitation we observe across all evaluated vision--language models, including ours. We further note that our task targets clip-level sub-action recognition rather than temporal action localisation: hand roles are assumed concurrent within a trimmed clip rather than temporally segmented. Future work will investigate adaptive frame sampling and temporal segmentation strategies to extend sub-action detection to longer, untrimmed egocentric sequences.

\clearpage

\section*{Acknowledgements}
This work was funded by the Leverhulme Trust Research Project Grant RPG-2023-079.
\bibliography{egbib}
\end{document}